\documentclass{llncs}

\usepackage{amsmath,amssymb,amsfonts,latexsym,graphicx,enumerate,setspace,blkarray, cite, color}
\usepackage{listings}
\begin{document}

\newcommand{\vv}[1]{\ensuremath{\mathbf{#1}}}
\newcommand{\uv}[1]{\ensuremath{\mathbf{\hat{#1}}}}
\newcommand{\rem}[1]{\ensuremath{\operatorname{rem}\text{ }#1}}
\newcommand{\aut}{\ensuremath{\operatorname{Aut}}}
\newcommand{\img}{\ensuremath{\operatorname{Img}}}
\newcommand{\im}{\ensuremath{\operatorname{Im}}}
\newcommand{\Frac}{\ensuremath{\operatorname{Frac}}}
\newcommand{\ord}{\ensuremath{\operatorname{ord}}}
\newcommand{\rank}{\ensuremath{\operatorname{rank}}}
\newcommand{\Gal}{\ensuremath{\operatorname{Gal}}}
\newcommand{\Int}{\ensuremath{\operatorname{Int}}}
\newcommand{\Div}{\ensuremath{\operatorname{Div}}}
\newcommand{\Var}{\ensuremath{\operatorname{Var}}}
\newcommand{\Id}{\ensuremath{\operatorname{Id}}}
\newcommand{\Spec}{\ensuremath{\operatorname{Spec}}}
\newcommand{\lcm}{\ensuremath{\operatorname{lcm}}}
\newcommand{\GL}{\ensuremath{\operatorname{GL}}}
\newcommand{\SL}{\ensuremath{\operatorname{SL}}}
\newcommand{\OL}{\ensuremath{\operatorname{O}}}
\newcommand{\normal}{\ensuremath{\triangleleft}}
\newcommand{\lamron}{\ensuremath{\triangleright}}
\newcommand{\one}{\ensuremath{\langle 1 \rangle}}
\newcommand{\zero}{\ensuremath{\langle 0 \rangle}}
\newcommand{\Q}{\ensuremath{\mathbb{Q}}}
\newcommand{\Z}{\ensuremath{\mathbb{Z}}}
\newcommand{\R}{\ensuremath{\mathbb{R}}}
\newcommand{\C}{\ensuremath{\mathbb{C}}}
\newcommand{\I}{\ensuremath{\mathbb{I}}}
\newcommand{\N}{\ensuremath{\mathbb{N}}}
\newcommand{\F}{\ensuremath{\mathbb{F}}}
\newcommand{\LC}{\ensuremath{\operatorname{LC}}}
\newcommand{\LT}{\ensuremath{\operatorname{LT}}}
\newcommand{\Hom}{\ensuremath{\operatorname{Hom}}}
\newcommand\spn{\operatorname{span}}
\newcommand\aff{\operatorname{aff}}
\newcommand\range{\operatorname{range}}
\newcommand\norm[1]{\left|\left| #1 \right|\right|}
\newcommand\inner[2]{\left< #1, #2 \right>}
\newcommand{\mset}[2]{\left\{ #1 \;\middle| \; #2 \right\}}

\title{Analytic Feature Selection for Support Vector Machines}
\author{Carly Stambaugh\inst{1}\and Hui Yang\inst{2} \and Felix Breuer\inst{1}\thanks{Felix Breuer was supported by Deutsche Forschungsgemeinschaft (DFG) grant BR~4251/1-1.}}
\institute{Department of Mathematics  \and Department of Computer Science \newline San Francisco State University\newline 1600 Holloway Avenue,\newline San Francisco, CA 94132\newline\email{ cstambau@mail.sfsu.edu},\email{ huiyang@sfsu.edu},\email{ felix@fbreuer.de}}
\maketitle
\begin{abstract}
Support vector machines (SVMs) rely on the inherent geometry of a data set to classify training data. Because of this, we believe SVMs are an excellent candidate to guide the development of an analytic feature selection algorithm, as opposed to the more commonly used heuristic methods. We propose a filter-based feature selection algorithm based on the inherent geometry of a feature set. Through observation, we identified six geometric properties that differ between optimal and suboptimal feature sets, and have statistically significant correlations to classifier performance. Our algorithm is based on logistic and linear regression models using these six geometric properties as predictor variables. The proposed algorithm achieves excellent results on high dimensional text data sets, with features that can be organized into a handful of feature types; for example, unigrams, bigrams or semantic structural features. We believe this algorithm is a novel and effective approach to solving the feature selection problem for linear SVMs.\end{abstract}
\begin{section}{Introduction}
Support Vector Machines (SVMs) are kernel-based machine learning classifiers\cite{Vapnik}. Using optimization methods such as quadratic programming, SVMs produce a hyperplane that separates data points into their respective categories. When a new, unlabeled, data point is introduced, its position relative to the hyperplane is used to predict the category the new point belongs to. One of the most important aspects of any machine learning classification problem is determining the particular combination of variables, or features, within a data set that will lead to the most accurate predictions, which is commonly known as the feature selection problem. Currently the methods used  by most machine learning engineers are heuristic in nature, and do not depend heavily on intrinsic properties of the data set\cite{HanKamber}. Due to the geometric nature of an SVM, it is natural to suggest that the performance of a particular feature set may be tied to its underlying geometric structure. This structure-performance relationship has in turn motivated us to develop an analytically driven approach to the feature selection problem for linear SVMs.
	
The primary goal of this research is to identify underlying geometric properties of optimal feature sets, and use these properties to create a feature selection algorithm that relies solely on the inherent geometry of a particular feature set. To accomplish this, we first create $n$-dimensional point clouds to represent known optimal and suboptimal feature sets. These point clouds are then used to identify structural differences between the optimal and suboptimal feature sets. Once these differences are identified, we design an algorithm to identify optimal feature sets based on these observations.

This feature selection algorithm is based on mathematical properties of the feature sets, making it analytic in nature. This sets the algorithm apart from the current, most widely used, wrapper-based or filter-based feature selection methods, which are mostly heuristic in nature\cite{HanKamber}. These methods sometimes require assumptions about the data set, for example, independence among the features, that might not be met by the data. Since our method is based on the geometric structure of the data set, it does not make such assumptions. Also, as machine learning techniques such as SVM become more widely adopted in various application domains, it is important to understand more about the interaction between a learner and a particular data set, as these insights may guide further development in the field. By discovering some mathematical properties that separate optimal feature sets from suboptimal feature sets, we can guide the feature selection process in a much more precise manner. Additionally, knowing these properties can help us to maximize the efficacy of SVMs for a particular classification problem. These properties could even be used to guide data collection efforts, in effect ensuring that the data collected is capable of providing a good feature space.

The algorithm is based on six properties that have been observed across several text data sets. The properties are based on dimensionality and intersection qualities of the affine hulls of the $n$-dimensional point clouds generated from a particular feature set. We evaluated the algorithm on several types of data sets, including low dimensional continuous data, low dimensional categorical data, high dimensional text data in a binary sparse vector format, and high dimensional text data in a word frequency-based sparse vector format. We identified the optimal feature sets of each data set using a wrapper based feature selection method which considers all possible subsets of the whole feature space. These optimal feature sets are then used to develop and evaluate the proposed feature selection algorithm, based on accuracy, precision and recall. We have observed that the algorithm delivers the  best performance on the high dimensional text data, in both binary and word frequency-based formats. The algorithm is best suited to data whose features can be grouped together into feature types, for example, unigrams and bigrams. 
	
Our algorithm achieves accuracies ranging from 76\% to 86\% within the data sets on which the model was trained, with an average precision of 86\%, and an average recall of 72\%. On test sets with dimensions ranging from 480 to 1440, accuracy ranges from 76\% to 86\%, with an average precision of 83\% and an average recall of 81\%. Precision remains high (.9-1) for data sets up to 3000 dimensions. However, the proposed algorithm does not perform well on test sets with dimension lower than approximately 500. More efforts are required to understand and address this phenomenon. While the CPU time used by the algorithm increases quadratically in both the number of features and the number of examples, the proposed algorithm requires no human interaction during its runtime. 

We believe that this algorithm has a significant impact on the problem of feature selection. Its analytic nature sets it apart from current, more heuristic, methods used widely throughout industry. The process requires no supervision from the user, and thus provides a marked reduction in man hours needed to determine optimal feature sets.
	 
\end{section}

\begin{section}{Related Work}
A great deal of studies have been carried out to identify the optimal features for a classification problem\cite{Molina}\cite{Joachims}. However, such studies are mostly heuristic in nature. In this section we review the two studies that are most germane to our proposed feature selection algorithm.

Garg, {\it et al.} introduce the projection profile; a data driven method for computing generalization bounds for a learning problem\cite{Garg}. This method is especially meaningful in high dimensional learning problems, such as natural language processing\cite{Garg}. The method hinges on random projection of the data into a lower dimensional space. Garg, {\it et al.} assert that if the data can be projected into a lower dimensional space with relatively small distortions in the distances between points, then the increase in classification error due to these distortions will be small\cite{Garg}. This is important, because in the lower dimension the generalization error bounds are smaller. Bradley, {\it et al.}\cite{Bradley} state that a lower data dimension also corresponds to a lower VC dimension, which in turn also causes lower generalization error\cite{Vapnik}. Expanding on this idea, we apply these concepts to the feature selection problem by quantifying a particular feature sets capacity for dimensionality reduction, giving preference to those feature sets that have the potential to produce lower generalization error.

Bern, {\it et al.} emphasize the importance of the maximizing the margin between reduced convex hulls in the case of non linearly separable data.\cite{Bern}. We investigate a relationship between classifier accuracy and properties of the intersection of the affine hulls of the class separated point clouds. In a sense, we are describing an upper bound on this margin, the idea being that the more intertwined the class separated point clouds are, the smaller the maximum margin between their reduced convex hulls becomes. We use the affine intersection of the class separated point clouds as a measure of a feature set's suitability for SVM classification with a linear kernel. The choice of affine hulls will be discussed further in the next section.
\end{section}

\begin{section}{Identifying the Relevant Geometric Properties of a Data Set}
\label{sec:geometry}

The overall approach of this work is to examine feature sets arising from  several natural language processing classification problems. We seek to identify key geometric properties that can be used to describe those feature sets for which an SVM performs well. This process of identifying relevant geometric properties is described in this section. In the next section, we construct an empirical algorithm for feature selection based on the geometric properties identified here.

Our training data consists of 717 feature sets, each manually labeled as optimal or suboptimal. These labels, based on classifier accuracy, were determined using an all subsets wrapper-based feature selection method on five data sets from four different classification problems. (These data sets are summarized in Table \ref{tab:trainData}, Section 5.) For each classification problem, we train every possible binary SVM using every possible subset of features. 

SVMs are inherently binary classifiers. There are several ways to address this when using SVMs for  multi class problems. A commonly used approach, as described in \cite{HanKamber}, is the {\em one vs all} approach. We use the following variation on this method. Consider a multi class classification problem with $\ell$ classes; this problem consists of $2^{\ell}$ classifiers that represent all possible ways to subdivide $\ell$ classes into two groups. We remove half of these possibilities due to symmetry. Finally, we do not consider the subset with an empty positive class to be a viable classifier, leaving $2^{\ell -1}-1$ possible binary classifiers. Each example in our training data represents one possible feature set for one possible binary classifier for a particular classification problem.

Because, for most of the five data sets we used, the number of samples we have is smaller than the total number of available features, we chose to focus on a linear kernel SVM, since, given the small number of samples, more complex kernels will likely lead to overfitting. As a linear kernel SVM performs linear separation if possible, it would have been natural to study the convex hulls of the positive and negative classes of samples. However, due to considerations of performance and ease of implementation, we instead chose to focus on a much simpler geometric invariant: the affine hulls of the positive and negative classes of samples. This choice allows us to use standard and widely-available linear algebra libraries for our computations so that we can work with high dimensional data sets, like those associated with natural language applications, in a manner that is computationally feasible.

In this paragraph, we review some basic material on affine hulls. For more information, we refer the reader to standard geometry textbooks such as \cite{webster,Ziegler1995}. Let $v_0,v_1,\dots, v_k$ be vectors. For any set of vectors, we write $(v_0,v_1,\dots,v_k)$ to denote the matrix with the $v_i$ as columns. The {\em linear hull} $\spn(v_0,v_1,\dots,v_k)$ of the vectors $v_i$ is the smallest linear space  containing all $v_i$ which, equivalently, can be defined as $\spn(v_0,v_1,\dots,v_k) = \mset{\sum \lambda_iv_i}{\lambda_i \in \R}$.
The dimension of the linear hull is the rank of the matrix with the $v_i$ as columns. An {\em affine space} is a translate of a linear space. In particular, it does not necessarily contain the origin. The {\em affine hull} $\aff(v_0,v_1,\dots,v_k)$ of the $v_i$ is the smallest affine space containing all $v_i$, which, equivalently, can be defined as
\begin{eqnarray*}
\aff(v_0,v_1,\dots,v_k) &=& \mset{\sum_{i=0}^k\lambda_iv_i}{\lambda_i\in \R, \sum_{i=0}^k\lambda_i = 1}.
% \\ &=& \mset{w +\sum_{i=0}^k\lambda_i(v_i-w)}{\lambda_i\in \R}
\end{eqnarray*}
The dimension of the affine hull is the dimension of the linear space that the affine hull is a translate of. In particular, the dimension of the affine hull of a point set is the same as the dimension of the polytope that is its convex hull. Thus, by definition, the dimension of the affine hull can be written as: 
\[
\dim(\aff(v_1,\ldots,v_d)) = \rank(v_1-v_d,\ldots,v_{d-1}-v_d)
\]
This simple observation makes calculations for higher dimensional data sets easy to implement and computationally efficient.

Now, suppose we have $n+1$ points in $n$-dimensional space. If the points are in general position, then the dimension of their affine hull is $n$. Moreover, assuming the points are in general position, then we can find a separating hyperplane for {\em any} partition of the points into two classes, i.e., the point set can be shattered. Somewhat surprisingly, it turns out however that the samples in our natural language processing data sets are not in general position. In fact, their affine hull has very low dimension, compared with the dimension of corresponding feature space. The ratio of the dimension of the affine hull and the dimension of the feature space in the data sets used to develop our training data are as low as .2, with an average of .52. Intuitively, if the ratio of the dimension of the affine hull over the dimension feature space is low, we expect the data set to contain a lot of structure, which the SVM can use to construct a classifier. (See also \cite{Garg,Bradley}, who show that if a data set can be effectively projected into a lower dimension with small distortions in the distances between points, the generalization error of that data set is lower than that of a data set lacking this property. ) This observation has led us to consider several geometric measures, called $f_1$ through $f_6$, defined in terms of simple ratios. We use these measures to assess the differences in the geometric structure of optimal and suboptimal feature sets with respect to the given data. 

Before we can define the properties $f_i$, we need to introduce some notation. The input data set to a binary classification problem is given in terms of a sparse matrix, with each point in the original data set represented as a row. The unique value of each feature is represented by a column in the matrix. That is, an entry $a_{ij}$ in the matrix is $1$ if the data point $i$ contains feature $j$ and it is $0$ otherwise. The rows of this matrix are organized into blocks such that each block contains all the data points belonging to the same class. We refer to this matrix as the full matrix, or $M_f$. The submatrix consisting of only the rows in the positive class is referred to as the positive matrix $M_p$ and the submatrix consisting of only the rows in the negative class is referred to as the negative matrix $M_n$. By considering every row as point in feature space, we can associate to each of these matrices a set of points in feature space.  We refer to the resulting three point sets as the {\em full point cloud}, $P_f$, the {\em positive point cloud}, $P_p$ and the {\em negative point cloud}, $P_n$. The {\em affine dimension} of a point cloud $P$ is the dimension of its affine hull, $\aff(P)$. To assess the dimension of the ambient space, we could use the dimension of feature space, i.e., the total number of columns. However, it may happen that some of the columns in a given matrix are zero, and as such columns contain no additional information, we chose to exclude them from our count. Therefore the {\em ambient dimension} is defined as the number of non-zero columns in a given matrix. Geometrically, this is the dimension of the smallest coordinate subspace the corresponding point cloud is contained in. Given this terminology, we now define ratios $f_1$ through $f_6$ as given in Table~\ref{tab:formulas}. Ratios $f_1$ through $f_5$ each contain in affine dimension in their numerator and an ambient dimension in their denominator. Ratio $f_6$ divides the number of samples contained in both of the affine hulls of $P_p$ and $P_n$ by the total number of samples.\footnote{Note that if, in the definition of $f_6$, we used the term convex hull instead of affine hull, a value of $f_6=0$ would guarantee linear separability. However, with our definition of $f_6$ a value of $f_6=0$ is neither necessary nor sufficient for separability.}

\begin{table}[ht]
\centering
\caption{Definitions of Geometric Properties $f_1$-$f_6$}
\begin{tabular}{l l r }
\hline \hline
property&numerator&denominator \\
\hline

$f_1$&affine dimension of $P_p$& ambient dimension of $P_p$\\
$f_2$&affine dimension of $P_n$& ambient dimension of $P_n$\\
$f_3$&affine dimension of $P_p$& ambient dimension of $P_f$\\
$f_4$&affine dimension of $P_n$& ambient dimension of $P_f$\\
$f_5$&affine dimension of $P_f$& ambient dimension of $P_f$\\
$f_6$&\# of samples in $\aff(P_p)\cap\aff(P_f)$\;\;\;\;\;   &\# of total samples\\
\hline
\end{tabular}
\label{tab:formulas}
\end{table}

The purpose of the properties $f_i$ is to allow us to assess the geometric structure of the data with respect to different feature sets. In this setting, a feature set is a set of columns. Selecting a certain subset of features amounts to removing the other columns from the matrices. Geometrically, this means projecting the point set onto the coordinate subspace corresponding to the selected feature set. We can then apply the measures $f_i$ to these projected data sets and compare the values we obtain.

For each feature set in our training data, we trained a linear kernel SVM on the training data and assessed the performance of the linear classifier obtained on the test data. We also computed the values of the $f_i$ for each feature set using the LinAlg library of NumPy\cite{numpy}.  
\begin{figure}[ht]
\begin{center}
\includegraphics[scale = .5]{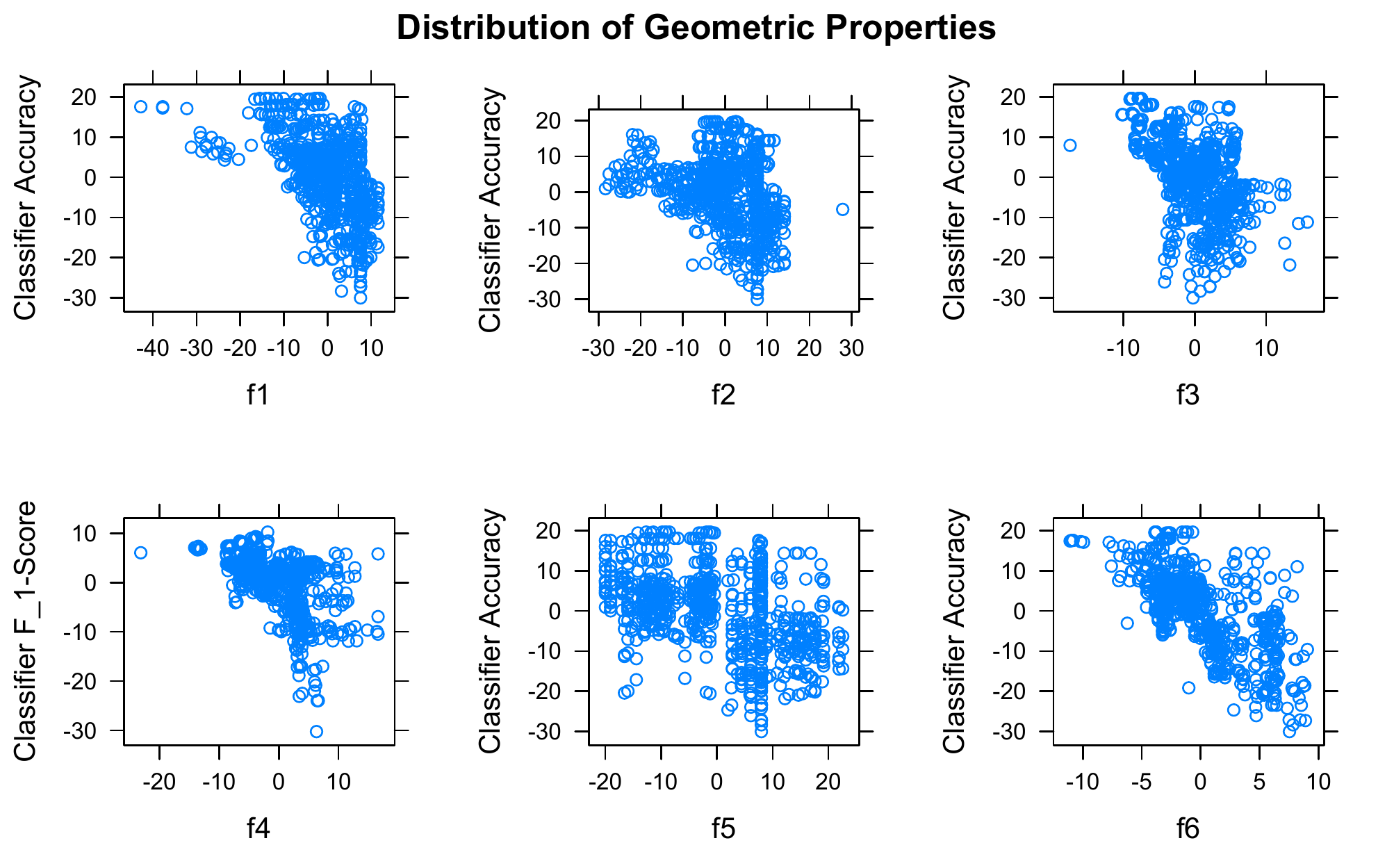}
\caption{Distributions of Geometric Properties}
\label{fig:AllProps}
\end{center}
\end{figure}

The results of this experiment are shown in Figure~\ref{fig:AllProps}. Each plot shows the standardized z-scores of the values of a particular geometric property for each of the 717 feature sets in our training data. The value of this ratio is plotted against a standardized measure of that particular feature set's performance. In most cases this measure is classifier accuracy, but in the case of $f_4$, we noticed a much stronger correlation between the $f_4$ value and the $F_1$-Score for a given feature set, which is defined as
$$\text{$F_1$-Score} = 2\cdot\frac{\text{precision}\cdot\text{recall}}{\text{precision}+\text{recall}}$$
where precision and recall are given by
\begin{eqnarray*}
\text{precision} = \frac{tp}{tp + fp} &\text{ and }&
\text{recall} = \frac{tp}{tp +fn}
\end{eqnarray*}
and $tp,fp,fn$ represent the number of true positives, false positives and false negatives, respectively. (These values are calculated by comparing the predicted values against the labels that were manually assigned during the generation of our training set.)
Notice the clear negative relationship between each of the properties and classifier performance. Each of the linear regression models pictured in Figure \ref{fig:AllProps} are significant on an $\alpha = .01$ level. 

Clearly, the geometric properties $f_i$ contain information about the quality of a given feature set. In the next section we use the $f_i$ as predictor variables to develop a logistic regression model, as well as a linear regression model that is the basis of our feature selection algorithm. We chose linear and logistic regression based on the observations in Figure \ref{fig:AllProps}, and the fact that we wish to determine whether a feature set is optimal or suboptimal, ultimately a binary decision. 
\end{section}

\begin{section}{Geometric Properties-Based Feature Selection Algorithm}
The goal of this algorithm is to use the observations discussed in the previous section to identify optimal feature sets for a text classification problem using an SVM with a linear kernel. This section describes the specifics of the algorithm.

The input includes a training data set, a list of categories used to label the data, a set of boundary values for the feature types and a directory to store the output files. The columns representing a given feature type must be stored in consecutive columns, as  previously described in Section 3. It is necessary for each training vector to contain at least one nonzero column for each feature type. If the data does not lend itself to this naturally, the user must handle missing values in the manner best suited to the particular learning problem. The vectors of the training data set should be represented in sparse vector notation.

\begin{figure}[h]
\caption{Structural Feature Selection Pseudo Code}
\begin{lstlisting}
1 for each unique binary classifier:
			
2   for each possible subset of features:
3	   generate training vectors for subset
4	   build positive, negative and full matrices
		
5	   for each matrix:
6		   calculate ambient and affine dimension
7	   calculate dimension based features:
		(see Section 3 for details)
8	   calculate affine intersection rate (f_6)
		(see Section 3 for details)
		
9 	standardize values for f1-f6 for all possible subsets
	
10  for each possible subset of features:
11	  lin_pred = predict using linear regression model
12	  log_pred = predict using logistic regression model

13	  if lin_pred > 0 and log_pred >= .5:
14		prediction = optimal
15		write subset to file
16	  else:
17		prediction = suboptimal
\end{lstlisting}
\label{fig:code}
\end{figure}
Figure \ref{fig:code} shows the structure of the algorithm. The program starts by identifying all the unique binary classifiers, and all the possible combinations of feature types(lines 1-2). It does this by generating all possible combinations of  labels and eliminates those which are symmetric to an existing subset. It is necessary to remove the empty feature set, and the binary classifier with an empty positive or negative class. The program creates a directory chosen by the user and creates files within it to store the results for each of the unique binary classifiers. Then, the program executes a nested loop as shown in figure \ref{fig:code}(lines 3-9). 
For each subset, we first need to process the training vectors so that they only include vectors for that particular feature set. Once this is done, the data points in the training set are split into positive and negative examples. Then, three matrices are used to represent the point clouds $P_n$,$P_p$ and $P_f$(line 5). The ratios, described in Section 3, are calculated using the affine and ambient dimensions of these point clouds.
\begin{table}[ht]
\centering
\caption{Logistic and Linear Regression Models}
\begin{tabular}{l r r }
\hline \hline
Predictor&Logistic Coefficient & Linear Coefficient \\
\hline
$\beta_0$&-0.64063267&-1.039011e-12\\
$f_1$&0.15706603&.0\\
$f_2$&0.1327297& 0\\
$f_3$&-0.03350878&09114375\\
$f_4$&-0.15182902& -.01223389\\
$f_5$&0.19548473&-.0200644\\
$f_6$&-0.68787718&0\\
\hline
\end{tabular}
\label{tab:models}
\end{table}

Finally, the algorithm makes a predication for a particular feature set based on the linear and logistic regression models detailed in Table \ref{tab:models}. These models were selected using forward stepwise inclusion with the AIC as the evaluation criterion. In order for a feature set to receive a prediction of {\em optimal}, the logistic regression model must predict a value greater than .5, and the linear model must predict a positive standardized accuracy. (Recall that a z-score of zero indicates the norm.) If both of these conditions are met, then the subset is written to the appropriate output file.
 
The output of the algorithm is a list of suggested feature sets that have the structural characteristics associated with optimal feature sets. Remember, an optimal subset need not be unique. The algorithm gives the user a list of subsets to chose from, based on the user's own criteria.

\end{section}

\begin{section}{Algorithm Evaluation}

In this section, we evaluate the power of the feature selection algorithm. We discuss some limitations of the algorithm, particularly, the relationship between the algorithm's performance and the dimensionality of the input data. We also present a theoretical and empirical time complexity analysis for the algorithm.

\begin{subsection}{Algorithm Performance}
The algorithm was run on each of the text data sets used to build the training set, and the results are presented in table \ref{tab:performance}. The polarity1, polarity2 and strength sentences are data sets originally used to classify the polarity and strength of relationships between a food/chemical/gene and a disease\cite{Yang}. The movies documents\cite{Movies} and webtext sentences\cite{nltk} are built from corpora included in Python's Natural Language Tool Kit\cite{nltk}. The movie review corpus is intended for classifying the polarity of reviews as positive or negative, and the webtext corpus consists of sentences from six different websites, each labeled according to their site of origin. 
\begin{table}[ht]
\centering
\caption{Summary of Data Suite Used to Train Model}
\begin{tabular}{l l l r r r}
\hline \hline
Data Set & R & C & BC & FT & resulting feature sets\\
\hline
Polarity1 Sentences &  463 & 645 & 7 & 5 & 156  \\
Polarity2 Sentences  &  324 & 600 & 7 & 5 & 183\\
Strength Sentences  & 787 & 645 & 7 & 5 & 179\\
Movies Documents   & 300 & 1000 & 1 & 3 & 7 \\
Webtext Sentences  & 1200 & 500 & 15 & 4 & 192\\
\hline
\end{tabular}
\label{tab:trainData}
\end{table}

Table \ref{tab:trainData} is a brief summary table of each set we used to train the model used in our feature selection algorithm. It includes the number of rows(R) and columns(C) of each raw data set. Each data set contains different types of features, and the number of these, (FT), is also listed for each data set. The number of unique binary classifiers (BC) resulting from the classification labels is also listed. Finally, the number of feature sets added to our training set as a result of the creation process is listed. 

To evaluate our feature selection algorithm, we calculate its accuracy, precision and recall by comparing the predictions made by the algorithm to the labels that were generated during creation of the training set. (See Section 3 for the label generation process.) Using these labels, we define accuracy, precision and recall as follows:
\begin{eqnarray*}
\text{accuracy} & = & \frac{tp + tn}{tp +fp + tn + fn}\\
\text{precision} &=& \frac{tp}{tp + fp}\\
\text{recall} &=& \frac{tp}{tp +fn},\end{eqnarray*} where $tp,tn,fp,fn$ represent the number of true positives, true negatives, false positives and false negatives, respectively. With respect to our algorithm, precision evaluates whether the feature sets selected by the algorithm actually perform optimally. Recall, on the other hand, measures how well the algorithm identifies all optimal feature sets. 
\begin{table}[ht]
\centering
\caption{Algorithm Performance for Training Data}
\begin{tabular}{l r r r}
\hline \hline
Data Set & Accuracy & Precision&  Recall\\
\hline
Polarity1 Sentences& $0.7564$ & $0.8621 $& $0.625 $\\
Polarity2 Sentences& $0.7542 $ & $0.6304 $& $0.8529$\\
Strength Sentences& $0.7814 $ & $0.8986 $& $0.6526 $\\
Movie Documents& $0.8571 $ & $1 $& $0.8 $\\
Webtext Sentences& $0.8091 $ & $0.9067 $& $0.6602 $\\
\hline
\end{tabular}
\label{tab:performance}
\end{table}
Recalling that an optimal feature set need not be unique, we see that precision is extremely important to this task. It is of more value to the user that the percentage of recommended feature sets that actually produce optimal results is high, since these results are the pool from which the user will ultimately choose a feature set. Optimal feature sets that are excluded from the results, or false negatives, do not have nearly as much consequence. 

Note, in table 4, the high precision within each data set. These numbers indicate that the algorithm we designed is quite effective for selecting optimal feature sets within the training data. Especially within the Movie Documents, where the algorithm achieves a precision of 1. This means that every feature set the algorithm returned was in fact an optimal feature set for classifying the Movies Documents with a linear SVM. While the algorithm's precision is somewhat lower on the Polarity2 Sentences, it is still impressive, given that only 38\% of the feature sets within the Polarity2 Sentences are actually labeled as optimal.

In the aforementioned data sets the full feature set is close to optimal, which means that running a linear SVM directly on the data with all features included gives almost the same accuracy as first running our feature selection algorithm and then applying the linear SVM. To assess if our algorithm can effectively reduce the dimension when the full feature set is not optimal, we ran the following experiment. The Polarity1 data set was modified by adding 25\% additional columns, increasing the total number of columns to 806. Each additional column was a random binary vector and received a random label. We applied our algorithm to each of the resulting binary classification problems. In all cases our algorithm recognized that the random columns did not contain relevant information and excluded them from the feature set. Applying the linear SVM to the reduced feature set, as selected by our wrapper algorithm, leads to a substantial improvement over applying the linear SVM directly to the full feature set: Accuracy increased by between 10\% and 26\% with a median increase of 15\%.

To test our algorithm on larger data sets, we created several data sets from the Amazon Customer Review Data, available from the UCI Machine Learning Repository\cite{UCI}. The raw data consists of 10,000 features and 1500 examples, with labels corresponding to 50 different authors. We developed each test set using a different set of five authors. Using different authors ensures that the reviews will be entirely different from one data set to the next. Because the reviews are different, the particular set of features generated will also be different, even though they are created in the same manner. The dimension of the  resulting data sets can increased or decreased by controlling the frequency requirements for inclusion of a feature. For example, to reduce the numbers of features, we would require that a particular unigram feature be present within the reviews at least 10 times. Then, to increase the dimension, we simply include less and less frequent features. Each test set also went through the same labeling process as the training data, in order to determine the algorithm's accuracy, precision and recall on previously unseen data. Recall this process was based on a wrapper based, all subsets algorithm that is commonly used to address the problem of feature selection. The results indicate that the algorithm also performs very well on previously unseen data. The Amazon data set was used to test the algorithm over a range of dimensions, and table \ref{tab:amazonRanges} summarizes the performance for these tests for column dimensions ranging from 480 to 1440. These results indicate that the algorithm performs very well within this range of column dimensions. We have observed that precision remains high (.9-1) for dimensions up to 3000.
\begin{table}[ht]
\centering
\caption{Peak Algorithm Performance for Amazon Data}
\begin{tabular}{l l l l}
\hline \hline
Dimension& Accuracy &Precision&  Recall\\
\hline
480	&0.768888889&	0.823529412&	0.711864407\\
640&	0.76&	0.838095238&	0.704\\
800&	0.831111111&	0.844444444&	0.870229008\\
960&	0.817777778&	0.837037037&	0.856060606\\
1120&	0.813333333&	0.822222222&	0.860465116\\
1280&	0.795555556&	0.8&	0.850393701\\
1440&	0.804444444&	0.82962963&	0.842105263\\
\hline
\end{tabular}
\label{tab:amazonRanges}
\end{table}
\end{subsection}

\begin{subsection}{Limitations}

As explained in Section~\ref{sec:geometry}, the proposed algorithm is designed to work well for linear kernel SVMs. In situations where the ratio of the number of samples to the total number of features is very large and the use of a higher degree kernel is warranted, we do not expect the affine geometry of the data set to reveal much useful information about which feature sets allow the SVM to generalize well. 

Moreover, the proposed algorithm is tailored towards binary data and we do not expect it to perform well on continuous data:  Suppose the data consists of $n$ points in $n$-dimensional space that are drawn from a model that generates points on a 1-dimensional affine subspace with a small additive error that is normally distributed. In this scenario the $n$ data points will span an affine space of dimension $n$, even though the true model is 1-dimensional. These theoretical considerations are confirmed by experiments which show that the algorithm does not perform well for continuous and categorical data. Table~\ref{tab:lowDimResults} provides a summary of the algorithm's performance on several test data sets according to column dimension and data type. A precision or recall score of 0 indicates that the algorithm did not accurately identify any optimal feature sets.
\begin{table}[ht]
\centering
\caption{Predictive Results for Low Dimensional Data}
\begin{tabular}{l l r r r}
\hline \hline
Column Dimension& Data Type& Accuracy & Precision&  Recall\\
\hline
13&\text{continuous}&0.3218& 0.1111& 0.2083\\
38&\text{categorical}&0.4444& 0 & 0\\
100&\text{categorical}&0.4286& 0 &0\\
\hline
\end{tabular}
\label{tab:lowDimResults}
\end{table}

Moreover, the data presented in Table~\ref{tab:lowDimResults} suggest that low dimensional data sets may limit the performance of the proposed algorithm. To better understand the relationship between our algorithm's performance and dimensionality, we designed an experiment using an Amazon data set as described above. The columns within each of the four feature types are organized in terms of frequency, so that the most common features occur in the earlier columns of each feature type block. The algorithm is used on these data sets repeatedly, while incrementing the number of dimensions included each time. For instance, the first run of the algorithm may include a total of 80 dimensions, the first 20 columns from each feature type. The algorithm's accuracy, precision and recall are recorded for the particular dimension, as well as the CPU time. The total number of features included is then increased to 160, by including the first 40 columns of each feature type. This process is repeated until all available dimensions are being used in the feature selection process. This is different than the previous Amazon data sets, because we are using the same set of five authors throughout the entire experiment, to control for variance between raw data sets. Figure 3 shows the results of this experiment. This experiment was repeated several times each using a different set of five authors with similar results.
\begin{figure}[h]
\begin{center}
\includegraphics[scale = .3]{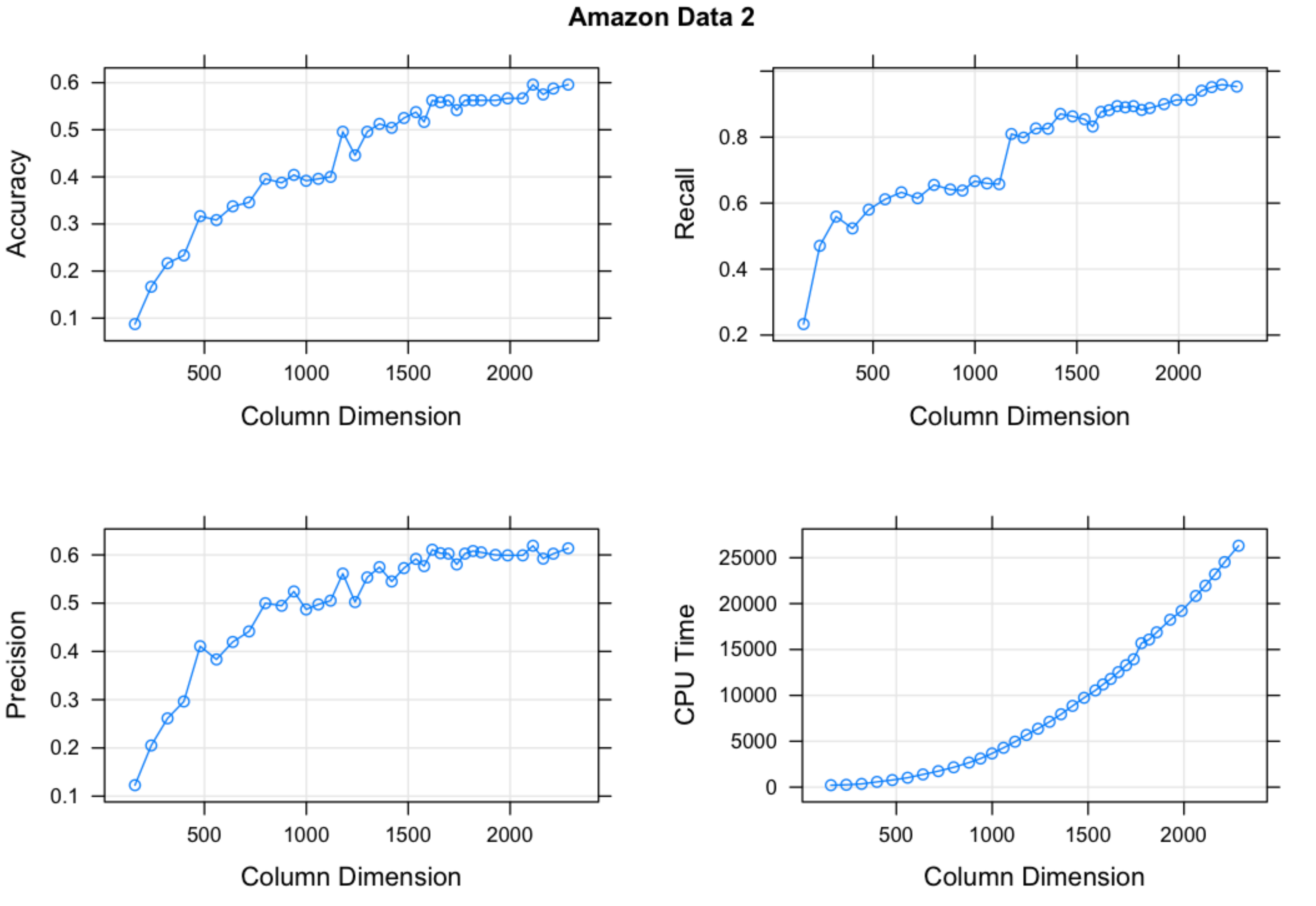}
\caption{Dimension Testing Results}
\end{center}
\label{fig:dimtests}
\end{figure}

These experiments indicate that the performance of the algorithm is very dependent on the dimensionality of the input data. Note the low values in accuracy, precision and recall for those data sets with less than ~400-500 columns. Figure 3 shows the rapid growth in accuracy, precision and recall for the lower dimensions that  becomes much slower for dimensions larger than ~500. Further study may be warranted to discover the cause of the dimensionality dependence observed in these experiments.

In figure 3, we see that the CPU time increases quadratically with column dimension. Note though, that the number of rows, feature types and labels are all held constant through out the experiment. The theoretical time complexity of the algorithm is in fact a function of all of these variables;
$$\mathcal{O}\left((2^{(\ell - 1)}-1)(2^k-1)(m^2n^2)\right),$$
where $\ell$ is the number of classification labels in the problem, $k$ is the number of feature groups present, and $m,n$ are the number of rows and columns, respectively, in the training data. The $\mathcal{O}(m^2n^2)$ terms come from the complexity of the singular value decomposition algorithm which is $\mathcal{O}(mn^2)$\cite{Tesic}. In our algorithm, we perform this calculation $(m + 2)$ times during the calculation of the affine intersection ratio. Recall, that the affine intersection ratio is calculated for $(2^k -1)$ feature sets, for each of $(2^{\ell -1} - 1)$ unique binary classifiers. While the Amazon data sets had the capacity to test up to 10,000 columns, the run time became unreasonably long after around 2400 dimensions on a lap top computer.
\end{subsection}

\end{section}

\begin{section}{Conclusion}
Support Vector Machines are machine learning classifiers that use geometry to separate training examples into their respective classes. Because of this, SVMs are an excellent candidate for a structural based feature selection algorithm. Many of the commonly used feature selection algorithms are heuristic in nature and do not use inherent characteristics of the data. A more data driven, analytic approach to feature selection will help machine learning engineers to better understand the relationship between a particular classification problem and a given optimal feature set. This understanding can influence data collection efforts and improve classification accuracy.

Through investigating the geometric structure of optimal and suboptimal feature sets, we found six qualities that differ significantly between them. We have discovered a linear relationship between the values of our dimensionality and intersection based features with classifier performance. We built linear and logistic regression models that use these six properties as predictor variables to identify optimal feature sets. We used these models to design a filter based feature selection algorithm that is analytic in nature, as opposed to the more commonly used wrapper based heuristic methods. 

Our feature selection algorithm performs best on text data sets that have more than approximately 500 features that can be organized into a handful feature types. While the precision remains high for data sets with more that ~2500 features, the computation time needed for these sets is too long to be practical on a single computer. Because of this, further study into parallelization of the algorithm may be warranted.

The algorithm did not perform well on low dimensional data sets. More study is needed to determine the cause of the relationship between the dimensionality of the original input data set. Currently, the algorithm does not support feature selection for SVMs using non linear kernels. However, we hypothesize that the algorithm could be successful when applied to other kernel types, if the data is first transformed using the chosen kernel, and the $f_i$'s are then calculated in the transform space. Further study is needed to accept or reject this hypothesis. Despite these limitations, our algorithm presents a useful and innovative approach to the problem of feature selection.
\end{section}

\bibliography{conference_paper}
\bibliographystyle{splncs}

\end{document}